\documentclass[10pt,conference]{IEEEtran}

\usepackage{algorithm}
\usepackage{algpseudocode}
\usepackage{subcaption}
\usepackage{diagbox}
\usepackage{tabularx}
\usepackage{comment}
\usepackage{makecell}
\usepackage{indentfirst}
\usepackage{booktabs,multirow}
\usepackage{bbm}
\usepackage{ulem}
\usepackage{textcomp}
\usepackage{xcolor}
\usepackage{cite}
\usepackage{amsmath,amssymb}
\usepackage{grffile}
\usepackage{graphicx}
\usepackage{dblfloatfix}
\usepackage{titlesec}
\usepackage{setspace}

\titlespacing*{\section}{0pt}{5pt plus 1pt minus 1pt}{2pt plus 1pt minus 1pt}
\titlespacing*{\subsection}{0pt}{4pt plus 1pt minus 1pt}{1pt plus 1pt minus 1pt}
\titlespacing*{\subsubsection}{0pt}{3pt plus 1pt minus 1pt}{1pt minus 0pt}

\IEEEoverridecommandlockouts
\def\BibTeX{{\rm B\kern-.05em{\sc i\kern-.025em b}\kern-.08em
		T\kern-.1667em\lower.7ex\hbox{E}\kern-.125emX}}

\begin{document}%%%%%%%%%%%%%%%%%%%%%%%%%%%%%%%%
% New definitions
\algnewcommand\algorithmicswitch{\textbf{switch}} 
\algnewcommand\algorithmiccase{\textbf{case}}
\algnewcommand\algorithmicassert{\texttt{assert}}
\algnewcommand\Assert[1]{\State \algorithmicassert(#1)}%
% New "environments"
\algdef{SE}[SWITCH]{Switch}{EndSwitch}[1]{\algorithmicswitch\ #1\ \algorithmicdo}{\algorithmicend\ \algorithmicswitch}%
\algdef{SE}[CASE]{Case}{EndCase}[1]{\algorithmiccase\ #1}{\algorithmicend\ \algorithmiccase}%
\algtext*{EndSwitch}%
\algtext*{EndCase}%
\flushbottom
% Allow display math (gather/align) to break across pages
\allowdisplaybreaks
% Discourage inline math at page-bottom line (IEEE format check)
\brokenpenalty=2000
\relpenalty=1000
\binoppenalty=1000
\emergencystretch=0.3em

\title{GSBF: Gaussian Splatting for \\Environment-Aware Beamforming\vspace{-0.3 cm}
}

\author{
	\IEEEauthorblockN{
		Yijie Bian, 
		Wei Guo, 
    Zixin Wang,\\
		Shenghui Song, \textit{Senior Member, IEEE},
		Jun Zhang, \textit{Fellow, IEEE}, and
		Khaled B. Letaief, \textit{Fellow, IEEE}
		} 
	% \IEEEauthorblockA{\IEEEauthorrefmark{1}Department of Electronic and Computer Engineering, \\The Hong Kong University of Science and Technology, Hong Kong, China} 
	\IEEEauthorblockA{Dept. of ECE, The Hong Kong University of Science and Technology, Hong Kong, China} 
	% \IEEEauthorblockA{\IEEEauthorrefmark{2}Frontiers Science Center for Mobile Information Communication and Security, \\Southeast University, Nanjing 210096, China\\}
	% \IEEEauthorblockA{\IEEEauthorrefmark{3}Key Laboratory of Measurement and Control of Complex Systems of Engineering, Ministry of Education, \\Southeast University, Nanjing 210096, China\\}
	% \IEEEauthorblockA{\IEEEauthorrefmark{4}School of Information Science and Engineering, Southeast University, Nanjing 210096, China}
\vspace{-0.45 cm}
} 
\maketitle

\bstctlcite{BSTcontrol}

\begin{abstract}
Beamforming plays a key role in multiple-input-multiple-output (MIMO) communication systems. However, conventional beamforming design normally requires accurate instantaneous channel state information (CSI) and iterative optimization, which incur substantial pilot overhead and computational complexity. Recognizing that radio propagation is intrinsically governed by the physical geometry, we develop a 3D Gaussian splatting for environment-aware beamforming (GSBF) pipeline based on multi-modal data, which characterizes the environment through a persistent 3D Gaussian representation. Specifically, GSBF models the environmental scattering response with reciprocity-preserving bidirectional spherical Gaussian (Bi-SG) kernels and performs two-sided electromagnetic rasterization to render an angular propagator map. The rendered map is then aggregated through an over-complete array-manifold dictionary and projected to the constant-modulus beamformers, thereby synthesizing beams directly from the access point (AP) pose and user position without online instantaneous CSI. Simulations demonstrate that GSBF consistently outperforms baselines such as exhaustive beam alignment (EBA) with lower latency.
\end{abstract}
% \vspace{-0.1 cm}

\section{Introduction}
\label{Introduction}
% The evolution toward prospective sixth-generation (6G) communications is driven by the demand for extreme data rates, ubiquitous coverage, and seamless connectivity in complex three-dimensional (3D) environments. To meet these requirements, advanced multi-antenna techniques, such as multiple-input multiple-output (MIMO) and highly directive beamforming, have become indispensable \cite{roadmap_6G}. However, with the increase of antenna arrays, the complexity of beamforming design also increases. Combined with rich-scattering environments, the ability to steer beams rapidly and accurately toward corresponding propagation paths determines the achievable spectral efficiency (SE) and the reliability of the wireless link.

Sixth-generation (6G) communications are envisioned to sustain seamless, high-capacity connectivity within spatially-complex wireless environments. While multiple-input-multiple-output (MIMO) and directive beamforming are essential to meet these demands \cite{roadmap_6G}, the expanding dimensionality of antenna arrays imposes a prohibitive computational burden on beamforming design. Hence, the ability to steer beams rapidly and accurately toward
corresponding propagation paths critically determines the achievable spectral efficiency (SE) and the reliability of the wireless link.

Conventional beamforming designs rely on explicit and accurate channel state information (CSI) obtained via sending dedicated pilots followed by iterative optimization \cite{HBFNet}, which incurs significant serial computational overhead. To address this issue, deep learning (DL) has been introduced to bypass online iterative overhead. Lin \textit{et al.} \cite{HBFNet} utilized DL frameworks to intelligently optimize hardware-constrained beamformers from practical channel estimation considering imperfect CSI. Meanwhile, Heng \textit{et al.} \cite{grid_less_beamforming} proposed a grid-free beamforming design using DL to directly generate beamforming vectors based on feedback from a few candidate beams. However, these methods still rely on designed pilots.

To alleviate the challenges, the channel knowledge map (CKM) was proposed to provide a new perspective of wireless channels from blind estimation to environment-awareness \cite{ckm_beam_alignment}. Wu \textit{et al.} \cite{ckm_beam_alignment} proposed two specific CKMs that map user position to channel relevant information to conduct training-free beam alignment. Furthermore, Shao \textit{et al.} \cite{enhanced_ckm_beam_alignment} proposed an enhanced CKM to recommend a variable beam search subspace for efficient beam alignment. However, CKM-based beamforming is hindered by the prohibitive overhead of site-specific wireless measurements and storage, while fixed codebook resolutions inevitably introduce performance loss.

Out-of-band information can also be used to bypass pilot-based CSI acquisition with sensors data. Le Magoarou \textit{et al.} \cite{location_beamforming} developed DL schemes that leverage user positions with embedding design for beamforming. To bridge the visual and wireless information, Ahn \textit{et al.} \cite{vision_beamforming} proposed computer-vision-aided beamforming frameworks using base station cameras and objects detection algorithms, but they remain sensitive to environmental conditions. Thus, multi-modal data, including light detection and ranging (LiDAR), radio detection and ranging (Radar), images and position information can be fused for mutual benefit \cite{environment_semantics_beam_alignment,lidar_gps_beam_tracking,sensing_aided_beam_tracking}. However, reliance on line-of-sight (LoS) sensors renders these methods prone to outages in non-line-of-sight (NLoS) scenarios. Moreover, since the implicit environment representations within neurons lack spatial causality and physical consistency, Bian \textit{et al.} \cite{VBS_beamforming,VBS_beamforming_journal} further proposed a virtual base station construction method for beam management to obtain the explicit and sparse representation from multi-modal data. However, due to the pure geometric-based framework, the coarse estimation and reconstructed beamspace are not accurate enough to support grid-free beamforming design.

To tackle the challenges and drawbacks mentioned above, this study proposes a novel Gaussian Splatting-based beamforming (GSBF) framework by utilizing multi-modal data, including LiDAR data and pose information. Specifically, this paper makes the following contributions: 1) We propose an online CSI-free and differentiable pipeline that deterministically maps a persistent 3D Gaussian scene representation to beamforming vectors; 2) within this pipeline, we design a bidirectional spherical Gaussian (Bi-SG) kernel that explicitly models the directional dependence of multi-path effects while enforcing physical reciprocity; and 3) we further propose a two-sided electromagnetic rasterization and synthesis strategy that facilitates the pipeline by aggregating complex-valued angular propagators into constant-modulus beamformers through an over-complete dictionary.

\section{System Model}
\label{system_model}
\begin{figure*}[t]
\centering
\includegraphics[width=0.999\textwidth]{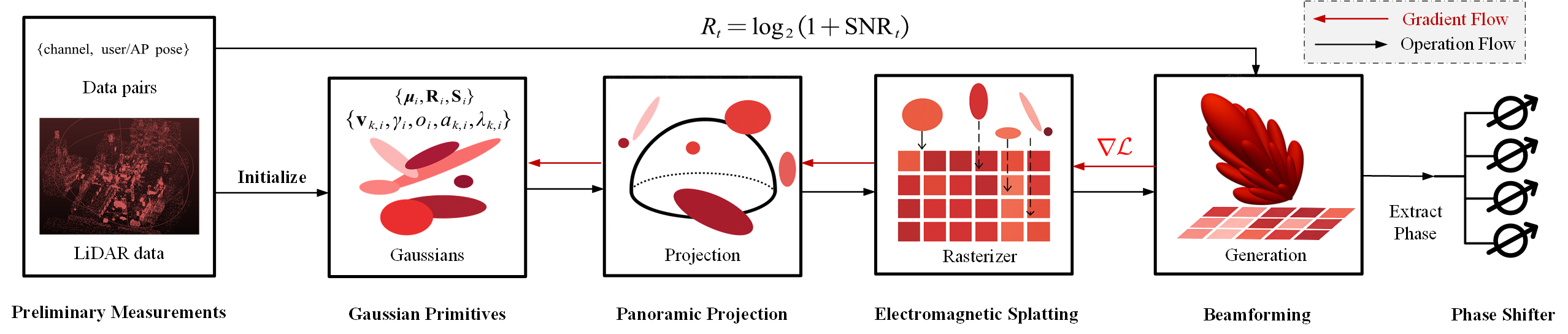}
\caption{Overview of the proposed GSBF framework, comprising four stages: (1) LiDAR-based initialization of Gaussian primitives; (2) panoramic projection onto the angular plane; (3) electromagnetic splatting for coherent propagator accumulation; and (4) synthesis of constant-modulus beamforming vectors. During training, the learnable Gaussian representation is optimized with CSI-related loss functions, while during inference, the optimized Gaussian representation enables direct beamforming generation from AP-user geometry without online CSI.}
\label{fig:GSBF_pipeline}
% \vspace{-0.6cm}
\end{figure*}
We consider an indoor scenario where the user is moving and the AP is static. The AP is equipped with a uniform planar array (UPA), with $N_\text{x}$ and $N_\text{y}$ elements along the x and y axes, respectively, giving a total of $N=N_\text{x}N_\text{y}$ antennas, and the user is equipped with a single antenna. The channel between the user and the AP at time slot $t$ is denoted as $\mathbf{h}_t\in \mathbb{C}^{1\times N}$. The AP applies a beamforming vector $\mathbf{f}_t\in \mathbb{C}^{N\times 1}$ to support the downlink communication. The received signal at user at time slot $t$ can then be expressed as
\begin{gather}
y_t = \sqrt{P_t}\mathbf{h}_t\mathbf{f}_tx_t + n_t,
\end{gather} 
where $ P_t $ denotes the transmit power, $x_t$ is the transmitted signal with zero mean and unit power, and $n_t\sim \mathcal{CN}(0,N_0\cdot\mathrm{BW})$ represents additive white Gaussian noise (AWGN) with noise power density $N_0$ and bandwidth $\mathrm{BW}$. The signal-to-noise ratio (SNR) at time slot $t$ is given by $ \text{SNR}_t = \frac{P_t|\mathbf{h}_t\mathbf{f}_t|^2}{N_0\cdot\mathrm{BW}} $. Thus, the SE at time slot $t$ can be calculated as
\begin{gather}
R_t = \log_2\left(1+\text{SNR}_t\right).
\end{gather}
We consider that the AP has a single radio frequency (RF) chain connected to an array of phase-shifters, thus the beamforming vector $\mathbf{f}_t$ has constant-modulus constraints. The objective is to design the beamforming vector $\mathbf{f}_t$ to maximize the SE $R_t$, which can be expressed as 
\begin{equation}
\begin{aligned}
\max_{\mathbf{f}_t} \quad & R_t, \\
\text{s.t.} \quad & \left|\{\mathbf{f}_t\}_i\right| = \frac{1}{\sqrt{N}}, \quad \forall i=1,2,\ldots,N.
\end{aligned}
\label{eq:objective}
\end{equation}
The constant-modulus constraint renders this problem non-convex, and iterative algorithms relying on online CSI incur prohibitive overhead for real-time applications.
% In the following sections, we propose a novel Gaussian splatting based beamforming (GSBF) framework to efficiently design the beamforming vector $\mathbf{f}_t$ given the parallel-processing capability of the 3DGS rasterization process and graphics processing unit (GPU) computation.

\section{Preliminaries on 3D Gaussian Splatting (3DGS)}
\label{sec:3dgs_prelim}
3DGS is a recently proposed explicit scene representation approach for high-quality and real-time novel view synthesis \cite{Kerbl3DGS}. It models a three-dimensional scene as a collection of explicit, anisotropic Gaussian primitives. Unlike implicit neural representations, 3DGS offers a persistent geometric backbone that is both differentiable and computationally efficient. Each Gaussian primitive is characterized by its spatial mean $\boldsymbol{\mu}_i \in \mathbb{R}^{3\times 1}$ and a volumetric covariance matrix $\boldsymbol{\Sigma}_i=\mathbf{R}_i \mathbf{S}_i \mathbf{S}_i^\top \mathbf{R}_i^\top \in \mathbb{R}^{3 \times 3}$, where $\mathbf{R}_i\in \mathbb{R}^{3 \times 3}$ is a rotation matrix and $\mathbf{S}_i\in \mathbb{R}^{3 \times 3}$ is a scaling matrix. The spatial influence of a primitive at a point $\mathbf{x}\in \mathbb{R}^{3\times 1}$ is defined by the exponential power distribution
\begin{equation}
    G(\mathbf{x}; \boldsymbol{\mu}_i, \boldsymbol{\Sigma}_i) = \exp \left( -\frac{1}{2} (\mathbf{x} - \boldsymbol{\mu}_i)^\top \boldsymbol{\Sigma}_i^{-1} (\mathbf{x} - \boldsymbol{\mu}_i) \right).
    \label{eq:3dgs}
\end{equation}

Unlike computationally intensive ray marching used in implicit neural rendering methods, the rendering process in 3DGS transforms these 3D primitives into a 2D image representation through a tile-based rasterization pipeline. Given a camera pose with position and orientation, the 3D Gaussian primitives are first projected onto a 2D image plane based on the camera parameters. The 2D Gaussian distribution $ G'(\mathbf{x}'; \boldsymbol{\mu}_i', \boldsymbol{\Sigma}_i') $ is then computed for each projected primitive with 2D mean $\boldsymbol{\mu}_i'\in \mathbb{R}^{2\times 1}$ and 2D covariance $\boldsymbol{\Sigma}_i'\in \mathbb{R}^{2 \times 2}$. To synthesize the final color $C$ for a pixel, the rasterizer sorts the corresponding $ M $ overlapping primitives by depth and accumulates their contributions using the front-to-back alpha-blending formula
\begin{equation}
    C = \sum_{i=1}^M c_i \alpha_i \prod_{j=1}^{i-1} (1 - \alpha_j)
\end{equation}
where $c_i$ is the color attribute derived from spherical harmonic functions and $\alpha_i=o_iG'(\mathbf{x}'; \boldsymbol{\mu}_i', \boldsymbol{\Sigma}_i')$ is the effective opacity of the $i$-th primitive, with $o_i$ being a learnable attenuation. Overall, this differentiable rasterization pipeline allows the geometric and radiometric attributes of the Gaussian primitives to be optimized, establishing a structured foundation for the more complex physical modeling in our proposed framework.

Recent 3DGS extensions to wireless radiation fields \cite{WRFGS_plus, RF3DGS, URF_GS} interpret rendered images as power angular spectra (PAS) or channel gain maps. However, these real-valued radiometric reconstructions cannot directly support complex-valued beamforming design, and their deformable architecture \cite{WRFGS_plus} compromises the physical interpretability of primitives that should remain persistent for static scattering objects. Moreover, their tight coupling between array dimensions and rendered PAS prevents environment-array decoupling. These drawbacks motivate our GSBF framework for direct beamforming generation from a persistent Gaussian representation.

\section{GSBF Framework}
\label{Proposed_GSBF}
% A1:
% In this section, we propose the GSBF framework to enable direct beamforming generation from environment-aware Gaussian representations. First, we introduce the Bi-SG representation and revisit the essence of beamforming. Then, we present the Gaussian projection in the antenna domain and the electromagnetic-aware rasterization. Furthermore, we describe the beamforming generation process and the constant-modulus projection.

In this section, we propose the GSBF framework to enable direct beamforming generation from environment-aware Gaussian representations. Specifically, we first review the foundation of utilizing 3DGS for beamforming, and then provide the details of the proposed GSBF framework.

\subsection{Explicit Beam Synthesis via Angular Propagators}
\label{sec:gsbf}
The synthesis of an analog beamforming vector $\mathbf{f}$ can be interpreted as the coherent reconstruction of a desired wavefront from the angular domain. Mathematically, $\mathbf{f}$ is a weighted superposition of steering vectors $\mathbf{a}_q \in \mathbb{C}^{N \times 1}$ spanning $Q$ directions \cite{essence_beamforming}, which can be calculated as 
\begin{gather}
\mathbf{f} = \sum_{q=1}^{Q} b_q \mathbf{a}_q,
\label{eq:essence_beamforming}
\end{gather}
where $b_q\in \mathbb{C}$ denotes the angular propagator in direction $q$. To ensure that the framework can capture sufficient spatial information, we employ an over-complete basis dictionary where $Q \ge N$. This oversampling preserves sufficient degrees of freedom to capture multi-path spatial features without rank deficiency. 
% Instead of compressing the environment into implicit neural parameters, GSBF utilizes Gaussian primitives as differentiable physical surrogates to ensure the generated propagators $ b_q $ are intrinsically consistent with the environment properties.

To bridge the environmental geometry with these angular propagators, our proposed GSBF utilizes Gaussian primitives as differentiable physical surrogates. Different from conventional beamforming methods that directly optimize a separate beamforming vector for each user position or channel realization, GSBF learns an environment-parameterized mapping from AP-user geometry to beamforming vectors. As shown in Fig. \ref{fig:GSBF_pipeline}, this mapping is realized by representing the environment with Gaussian primitives equipped with Bi-SG kernels, projecting these primitives onto the antenna-domain angular plane, and rasterizing their electromagnetic contributions into an angular propagator map. Under such explicit representation, we can generate the angular propagators $ \{b_q\} $, that are intrinsically consistent with the environment. The beamforming vector is then synthesized by aggregating the rendered propagators through an over-complete dictionary of steering vectors, followed by a projection onto the constant-modulus set to satisfy hardware constraints.

% As shown in Fig. \ref{fig:GSBF_pipeline}, we design a Bi-SG kernel for each Gaussian primitive to represent the electromagnetic response of the scattering environment. Subsequently, we employ an panoramic projection to project the 3D Gaussian primitives onto the antenna receiving plane, and then crop the result to obtain the angular propagator map corresponding to the antenna's forward hemisphere. 

% Under such explicit representation, we can generate the angular propagators $ b_q, q=1,\cdots,Q $, that are intrinsically consistent with the physical environment. Finally, the beamforming vector is synthesized by aggregating the rendered propagators through the over-complete dictionary of steering vectors, followed by a projection to the constant-modulus set to satisfy the hardware constraints.   

\subsection{Bi-SG Kernel Design}
\begin{figure}
  \centering
  \includegraphics[width=0.45\textwidth]{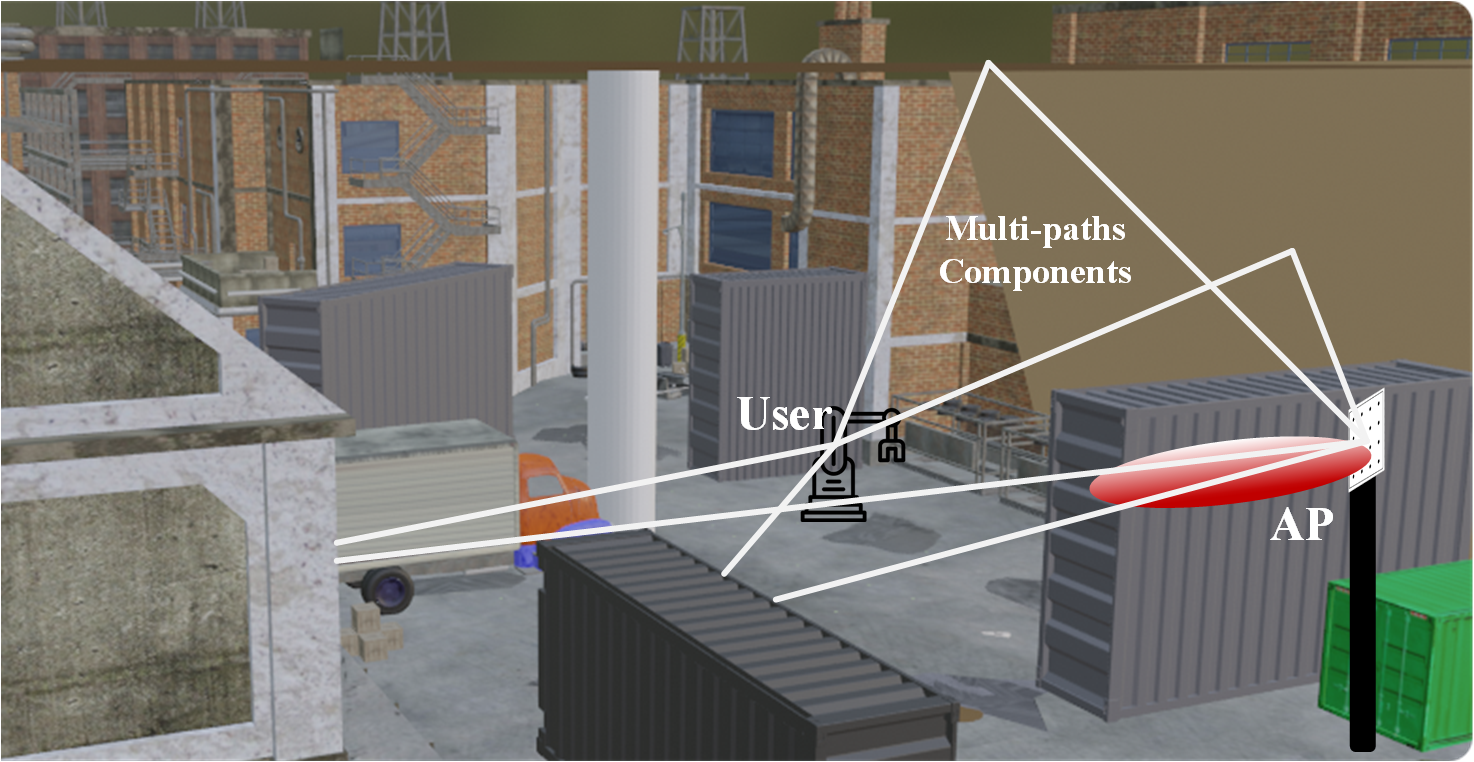}
  \caption{The synthetic factory scene constructed in Blender.}
  \label{fig:scene}
\end{figure}
% While the 3D Gaussian distribution characterizes the spatial presence and boundary of a primitive, the Bi-SG kernel serves as its electromagnetic counterpart, explicitly modeling the directional scattering response as a function of propagation geometry. The objectives of designing the kernel include two main aspects: 1) to effectively capture the physical scattering characteristics of the environment while retaining the interpretable nature; 2) to decouple the environment from the intrinsic properties of the antenna array as much as possible. To achieve these goals, we design a Bi-SG kernel using Gaussian spherical functions and bidirectional modeling motivated by \cite{SG_splatting,BiGS}. Bidirectional architecture can model both the incident and reflected wavefronts associated with each Gaussian primitive, while spherical Gaussian functions can provide a compact and efficient high-frequency aware representation of the angular scattering patterns.

While standard 3D Gaussian distributions characterize the spatial occupancy and geometric boundaries of scene primitives, they lack the intrinsic mechanisms to describe complex electromagnetic interactions. In wireless beamforming, the scene representation should characterize how each primitive responds to electromagnetic waves arriving from and scattering toward different directions. We therefore augment each Gaussian primitive with a learnable kernel, whose design is guided by two objectives: 1) to effectively capture the physical scattering characteristics of the environment while maintaining interpretability; and 2) to decouple environmental scattering properties from the intrinsic response of the antenna array. To achieve these objectives, we tailor a Bi-SG kernel using spherical Gaussian functions and a bidirectional parameterization motivated by \cite{SG_splatting,BiGS}. The bidirectional parameterization captures the coupled geometry between incident and scattered wavefronts, while the spherical Gaussian lobes provide a compact and efficient representation of angular scattering patterns.

For a signal incident from direction $\boldsymbol{\omega}_{\text{in}}$ and scattering towards direction $\boldsymbol{\omega}_{\text{out}}$, the Bi-SG kernel parameterizes the scattering response of the $i$-th Gaussian primitive $S_i$ as
\begin{gather} 
S_i(\boldsymbol{\omega}_{\text{in}}, \boldsymbol{\omega}_{\text{out}}) = 
\gamma_i + \sum_{k=1}^{K} a_{k,i} \exp[ \lambda_{k,i} (\kappa_{k,i} - 1) ],
\label{eq:bi_sg_kernel} 
\end{gather}
where $\gamma_i \in \mathbb{C}$ and $a_{k,i} \in \mathbb{C}$ are the learnable complex-valued albedo and lobe amplitudes, $\mathbf{v}_{k,i} \in \mathbb{R}^{3\times 1}$ is the unit-normalized learnable axis of the $k$-th scattering lobe, and $\lambda_{k,i} \in \mathbb{R}^+$ denotes its sharpness. To smoothly adapt the directional response across different propagation geometries, the alignment metric $\kappa_{k,i} = \beta(\boldsymbol{\omega}_{\text{in}}^\top \mathbf{v}_{k,i})(\boldsymbol{\omega}_{\text{out}}^\top \mathbf{v}_{k,i}) + (1-\beta)\mathbf{e}^\top\mathbf{v}_{k,i}$ blends the near-collinear propagation and redirected-scattering alignment terms, where $\beta=\sigma[\nu(\boldsymbol{\omega}_{\text{in}}^\top\boldsymbol{\omega}_{\text{out}}-\xi)] \in (0,1)$ is the collinearity-based blending weight, $\sigma(x)=1/(1+\exp(-x))$ is the sigmoid function, $\nu>0$ controls the transition steepness, and $\xi\in(-1,1)$ specifies the collinearity threshold. Moreover, $\mathbf{e}=(\boldsymbol{\omega}_{\text{in}}-\boldsymbol{\omega}_{\text{out}})/\|\boldsymbol{\omega}_{\text{in}}-\boldsymbol{\omega}_{\text{out}}\|_2$ represents the scattering half-vector-related direction for non-collinear propagation.

Unlike traditional unidirectional representations, this formulation in \eqref{eq:bi_sg_kernel} ensures $S_i(\boldsymbol{\omega}_{\text{in}},\boldsymbol{\omega}_{\text{out}})=S_i(-\boldsymbol{\omega}_{\text{out}},-\boldsymbol{\omega}_{\text{in}})$ under link reversal. This intrinsic symmetry ensures that the learned environment representation is consistent with reciprocal wave propagation under the adopted direction convention.

\subsection{Panoramic Projection}
Unlike the vanilla 3DGS, which projects the primitives based on the pinhole camera model \cite{Kerbl3DGS}, the received signal at AP-side is a superposition of the contributions from all directions in the forward hemisphere. Thus, our framework adopts an equirectangular projection to first project the 3D Gaussian primitives onto a panoramic angular map with $ W\times H $ resolution, and then crop the result to obtain the angular propagator map corresponding to the antenna's forward hemisphere. 

To be specific, let $\mathbf{p}_\text{w} = [p_{\text{wx}}, p_{\text{wy}}, p_{\text{wz}}, 1]^\top$ denote the Gaussian primitive center in the world coordinate system. Then, with the viewing matrix\footnote{The viewing matrix $\mathbf{V} = \begin{bmatrix} \mathbf{W} & \mathbf{t} \\ \mathbf{0}^\top & 1 \end{bmatrix} \in \mathbb{R}^{4\times 4}$ is a homogeneous transformation matrix that defines the rigid-body transformation from the world space to the local coordinate system of the observer, where $\mathbf{W} \in \mathbb{R}^{3\times 3}$ is the rotation matrix and $\mathbf{t} \in \mathbb{R}^3$ is the translation vector.} $\mathbf{V}$, we transform it to the local coordinate system as $ \mathbf{p} = \mathbf{V} \mathbf{p}_\text{w} = [p_\text{x}, p_\text{y}, p_\text{z}, 1]^\top $. The azimuth $\phi$ and elevation $\theta$ are then mapped to pixel coordinates $\mathbf{p}'=[u, v]^\top$ via $u = W(\frac{\phi}{2\pi}+\frac{1}{2})$ and $v = H(\frac{\theta}{\pi}+\frac{1}{2})$, where $ \theta= \arctan2(p_{\text{y}}, \rho), \phi = \arctan2(p_{\text{x}}, p_{\text{z}}), \rho = \sqrt{p_{\text{x}}^2 + p_{\text{z}}^2}$. The 3D covariance $\mathbf{\Sigma}_i$ in world space is then propagated to the 2D plane by
\begin{gather}
\mathbf{\Sigma}_i' = \mathbf{J}_i\,\mathbf{W}\,\mathbf{\Sigma}_i\,\mathbf{W}^\top\mathbf{J}_i^\top,
\label{eq:LocalCovariance}
\end{gather}
where $\mathbf{J}_i$ is the Jacobian matrix. The elements of $\mathbf{J}_i$ are given by
\begin{gather}
\mathbf{J}_i = \begin{bmatrix}
 \dfrac{W}{2\pi} \dfrac{p_{\text{z}}}{\rho^2} & 0 & -\dfrac{W}{2\pi} \dfrac{p_{\text{x}}}{\rho^2} \\
 -\dfrac{H}{\pi} \dfrac{p_{\text{x}}p_{\text{y}}}{\rho\,r^2} & \dfrac{H}{\pi} \dfrac{\rho}{r^2} & -\dfrac{H}{\pi} \dfrac{p_{\text{z}}p_{\text{y}}}{\rho\,r^2}
\end{bmatrix},
\label{eq:Jacobian}
\end{gather}
where $ r = \sqrt{p_{\text{x}}^2 + p_{\text{y}}^2 + p_{\text{z}}^2} $. This panoramic projection strategy ensures the spatial causality of the environment is preserved in the angular domain, which is consistent with antenna receive characteristics.

\subsection{Electromagnetic Rasterization }
Based on the projected positions and covariances defined above, electromagnetic rasterization assigns each projected Gaussian primitive an electromagnetic contribution and accumulates these contributions into an angular-domain propagator map. This process is divided into a source-to-primitive transmittance computation and a coherent propagator accumulation at AP-side. Given the reciprocity assumption of the wireless channel, we treat the user as the virtual source and the AP as the observer during rasterization. In the following subsections, we will describe the two steps in detail.

\subsubsection{Source-to-Primitive Transmittance }
To accurately characterize the spatial causality of the wireless channel, we first account for the electromagnetic shadowing effects between the user and each environment primitive $i$. We introduce a source-side transmittance factor $\tau_i^\text{UE} $, which quantifies the residual signal reaching the $i$-th Gaussian primitive. 

For each Gaussian primitive $ i $, the rasterizer evaluates the attenuation contributed by preceding primitives $ j $ from the perspective of the source. Instead of sorting by depth, the sequence of preceding primitives is determined by their spatial distance relative to the source. The transmittance $\tau_i^\text{UE}$ is computed through a front-to-back cumulative product,
\begin{equation}
    \tau_i^\text{UE} = \prod_{j \in \mathcal{R}_i} (1 - \delta_{ij}^{\text{UE}}),
\end{equation}
where $\mathcal{R}_i$ denotes the set of occluder primitives that are located between the user and the target primitive $i$, $\delta_{ij}^{\text{UE}}= o_j G'(\mathbf{p}_i'; \mathbf{p}_j', {\mathbf{\Sigma}_j^\text{UE}}')$ represents the attenuation contribution of occluder $j$ at the projected center of primitive $i$ in user-side projection, and $ \mathbf{p}_i' $ is the projected position of Gaussian primitive $i$ in user-side projection. The rotation matrix $\mathbf{W}_i^{\mathrm{UE}}$ is dynamically constructed for each target primitive $i$ such that the local viewing axis is aligned with the direction from the user to the primitive center $\boldsymbol{\mu}_i$, and ${\boldsymbol{\Sigma}_j^{\mathrm{UE}}}'$ is then computed by \eqref{eq:LocalCovariance}. The source-to-primitive transmittance represents the signal received by each Gaussian primitive after accounting for the attenuation effects of the environment, which serves the subsequent AP-side rasterization and synthesis.

\subsubsection{Coherent Propagator Accumulation }
Following the determination of $\tau_i^\text{UE}$, the rasterizer synthesizes the final propagator map by projecting scattering signatures onto the angular plane of the static AP. The complex-valued contribution of primitive $i$ to an angular pixel at coordinates $(u, v)$ is defined as 
\begin{equation}
    b_i(u, v) = o_i\tau_i^\text{UE}\,\tau_i^\text{AP}(u,v)\,S_i(\boldsymbol{\omega}_\text{in}, \boldsymbol{\omega}_\text{out})\,G'(\mathbf{q}'; \mathbf{q}_i', {\mathbf{\Sigma}_i^\text{AP}}'),
\end{equation}
where $\tau_i^\text{AP}(u,v)$ denotes the AP-side transmittance at pixel $(u,v)$, $\mathbf{q}'$ denotes the position of the pixel in the AP-plane, and $\mathbf{q}_i'$ denotes the projected AP-plane position of primitive $i$.  The incident direction $ \boldsymbol{\omega}_{\text{in}} $ and scattering direction $ \boldsymbol{\omega}_{\text{out}} $ for primitive $i$ are determined by the relative geometry between the Gaussian primitive center $ \boldsymbol{\mu}_i $, the user position, and the AP position, respectively. $ S_i(\boldsymbol{\omega}_\text{in}, \boldsymbol{\omega}_\text{out}) $ can then be computed according to \eqref{eq:bi_sg_kernel}. The AP-side transmittance can be similarly computed using the attenuation of overlapping primitives,
\begin{equation}
    \tau_i^\text{AP}(u,v) = \prod_{j \in \mathcal{T}_{i,u,v}} \left(1 - \delta_{ij}^{\text{AP}}(u,v)\right),
\end{equation}
where $\mathcal{T}_{i,u,v}$ is the set of Gaussian primitives whose AP-side projections overlap pixel $(u,v)$. The effective AP-side attenuation contribution $ \delta_{ij}^{\text{AP}}(u,v) = o_j G'(\mathbf{q}'; \mathbf{q}_j', {\mathbf{\Sigma}_j^\text{AP}}') $ of occluder $j$ can be similarly computed. The rotation matrix $\mathbf{W}^{\mathrm{AP}}$ is the rotation component of the fixed AP extrinsic matrix, and $ {\mathbf{\Sigma}_j^\text{AP}}' $ is similarly computed by \eqref{eq:LocalCovariance}. 

The aggregate complex-valued angular propagator $b_{u, v}$ at AP-side can then be obtained by summing the weighted contributions of all overlapping primitives $\mathcal{S}_{u,v}$,
\begin{equation}
    b_{u, v} = \sum_{i \in \mathcal{S}_{u,v}} b_i(u, v).
\end{equation} 
The propagator matrix $\mathbf{B}_t = \{b_{u, v}\} \in \mathbb{C}^{H \times W^\prime}$, where $ W^\prime=\frac{1}{2}W, u\in\{\frac{1}{4}W+1, \dotsm, \frac{3}{4}W-1, \frac{3}{4}W\} $, selects the forward-hemisphere angular space. The coherent propagator accumulation synthesizes the complex-valued map at AP-side by superimposing the contributions of all related Gaussian primitives, treating them as the secondary sources of the electromagnetic field.

\subsection{Beamforming Generation}
In the following process, we map the rendered angular propagator map to antenna-domain excitations and enforce the analog constant-modulus constraints required by the hardware, to ensure that the synthesized vector follows the physical constraints of the UPA array.

As mentioned in Section \ref{sec:gsbf}, the number of steerable directions in the angular domain must be oversampled to capture the detailed features in the wireless environment. To achieve this, the middle half of the rendered angular panorama, corresponding to the antenna's forward hemisphere, is sampled on a high-resolution grid $\mathbf{A} = [\mathbf{a}_{1,1}, \dots, \mathbf{a}_{i,j}, \dots, \mathbf{a}_{H,W^\prime}]\in \mathbb{C}^{N \times Q}$, where $W^\prime=W/2$ and $Q=HW^\prime$. The steering vector $\mathbf{a}_{i,j}$ corresponding to direction $(\theta_i,\phi_j)$ is defined as
\begin{equation}
    \mathbf{a}_{i,j} = \frac{1}{\sqrt{N}} \left[ e^{\mathrm{j} \Psi_{0,0}}, \dots, e^{\mathrm{j} \Psi_{m,n}}, \dots, e^{\mathrm{j} \Psi_{N_\text{x}-1,N_\text{y}-1}} \right]^\top,
\end{equation}
where $\Psi_{m,n}=m\Phi_\text{x}+n\Phi_\text{y}$ represents the phase shift of the antenna element located at the $m$-th column and $n$-th row. The pixel-center angular samples are given by $\theta_i=\pi[(i-\frac{1}{2})/H-\frac{1}{2}]$ and $\phi_j=\pi[(j-\frac{1}{2})/W^\prime-\frac{1}{2}]$. Assuming half-wavelength antenna spacing, the spatial phase increments along the x and y axes are $\Phi_\text{x}=\pi\cos\theta_i\sin\phi_j$ and $\Phi_\text{y}=\pi\sin\theta_i$, respectively, where $m=0,\ldots,N_\text{x}-1$ and $n=0,\ldots,N_\text{y}-1$ denote the antenna-element indices.

The angular propagator map rendered by GSBF is first vectorized into $\mathbf{b}_t = \mathrm{vec}(\mathbf{B}_t)$. Then, the synthesis of the unnormalized beamforming vector $\tilde{\mathbf{f}}_t$ is formulated as a linear aggregation based on the foundation in \eqref{eq:essence_beamforming},
\begin{equation}
    \tilde{\mathbf{f}}_t = \mathbf{A} \mathbf{b}_t.
\end{equation}
% This operation transforms the explicit geometry representation into an antenna-domain excitation and separates the environment and array response, where each element in $\tilde{\mathbf{f}}_t$ represents the coherent superposition of all scattering contributions in the scene. 
Given the constraints of a fully-analog architecture, the beamforming vector must satisfy the constant-modulus requirement for each antenna element. We project $\tilde{\mathbf{f}}_t$ onto the feasible domain to obtain the final beamforming vector $\hat{\mathbf{f}}_t$ as follows,
\begin{equation}
	\{\hat{\mathbf{f}}_t\}_i = \dfrac{\{\tilde{\mathbf{f}}_t\}_i}{\sqrt{N}|\{\tilde{\mathbf{f}}_t\}_i|}.
  \label{eq:normalization_bf}
\end{equation}

Since the beamforming vector is generated by the environment-parameterized forward path, SE maximization is performed by optimizing the learnable representation parameters over offline training samples, after which the learned parameters are reused for online beam synthesis without per-slot beamformer optimization. Specifically, we minimize $\mathcal{L}=D_t+\lambda_\text{cm}L_\text{cm}$, where $D_t=1-\frac{|\mathbf{h}_t\hat{\mathbf{f}}_t|^2}{||\mathbf{h}_t||_2^2} $ represents the normalized beamforming power-gain gap. The regularization term $L_\text{cm} = \sum_{i=1}^{N}\left(\left|\{\tilde{\mathbf{f}}_t\}_i\right| - \mu\right)^2 / \mu^2$ penalizes the amplitude variation of the unnormalized beamforming vector before projection with regularization weight $\lambda_\text{cm}$, where $\mu = \frac{1}{N} \sum_{i=1}^{N} \left|\{\tilde{\mathbf{f}}_t\}_i\right|$. The optimized parameters include the Gaussian spatial parameters $\{\boldsymbol{\mu}_i, \mathbf{R}_i, \mathbf{S}_i\}$ and the Bi-SG kernel parameters $\{\mathbf{v}_{k,i}, \gamma_i, o_i, a_{k,i}, \lambda_{k,i}\}$, together with density control \cite{Kerbl3DGS}. To mitigate the gradient mismatch introduced by the constant-modulus projection in \eqref{eq:normalization_bf}, we adopt the straight-through estimator (STE) during gradient backward propagation.

% Since our objective is to maximize the SE in \eqref{eq:objective}, we define the loss function as $ \mathcal{L}=D_t+\lambda_\text{cm}L_\text{cm}$. $ D_t $ is the difference between the SE of the digital upper bound and the current training SE. $ L_\text{cm} = \sum_{i=1}^{N}\left(\left|\{\tilde{\mathbf{f}}_t\}_i\right| - \mu\right)^2 / \mu^2 $, where $ \mu = \frac{1}{N} \sum_{i=1}^{N} \left|\{\tilde{\mathbf{f}}_t\}_i\right| $ is the constant-modulus regularization term and $ \operatorname{var}(\cdot) $ denotes the variance of the amplitude. The optimization process is to adjust the learnable parameters, including the distribution related parameters $ \{\boldsymbol{\mu}_i, \mathbf{R}_i, \mathbf{S}_i\} $ and kernel related paramters $ \{\mathbf{v}_{k,i}, \gamma_i, o_i, a_{k, i}, \lambda_{k, i}\} $ with density control \cite{Kerbl3DGS}. To mitigate the gradient vanishing issue caused by the projection, we adopt the straight-through estimator (STE) for \eqref{eq:normalization_bf} during gradient backward propagation. 

% This encourages the unnormalized beamforming vector $\tilde{\mathbf{f}}_t$ to be close to the constant-modulus constraint, and $\lambda_\text{cm}$ is the corresponding regularization weight.

\section{Simulation Setup and Results}
\subsection{Simulation Setup}
We create a synthetic dataset based on the factory scene constructed from open-source components shown in Fig. \ref{fig:scene}, to validate GSBF using a two-stage pipeline: scene construction in Blender and ray-tracing in Sionna. The central frequency $ f_\text{c} $ is set to 28 GHz, with bandwidth $ \mathrm{BW} =100 \text{~MHz} $, noise power density $ N_0 = -174 \text{~dBm/Hz}$, and transmit power $ P_\text{T} = 10 \text{~dBm} $. During training, the model is provided with regional LiDAR data, the AP pose, the user position, and the corresponding channel. The LiDAR data is used to initialize the Gaussian primitives. At the evaluation stage, only the AP pose and user position are required. The user position is randomly sampled from a $ 20 \times 15 \text{~m}^2 $ area. The rendered angular propagator map has a resolution of $ H=W^\prime=45 $ with over-sampled angular pixels to ensure a sufficiently dense representation for beam synthesis. The number of lobes $K$ is set to 4, $\nu$ and $\xi$ are set to 80 and 0.95, respectively, and $\lambda_{\mathrm{cm}}$ is set to 0.1. The GSBF model is trained on 5600 samples and tested on 1400 samples.

\subsection{Results and Discussion}
\begin{figure*}[t!]
  \centering
  \begin{subfigure}{0.323\textwidth}
    \centering
    \includegraphics[width=\linewidth]{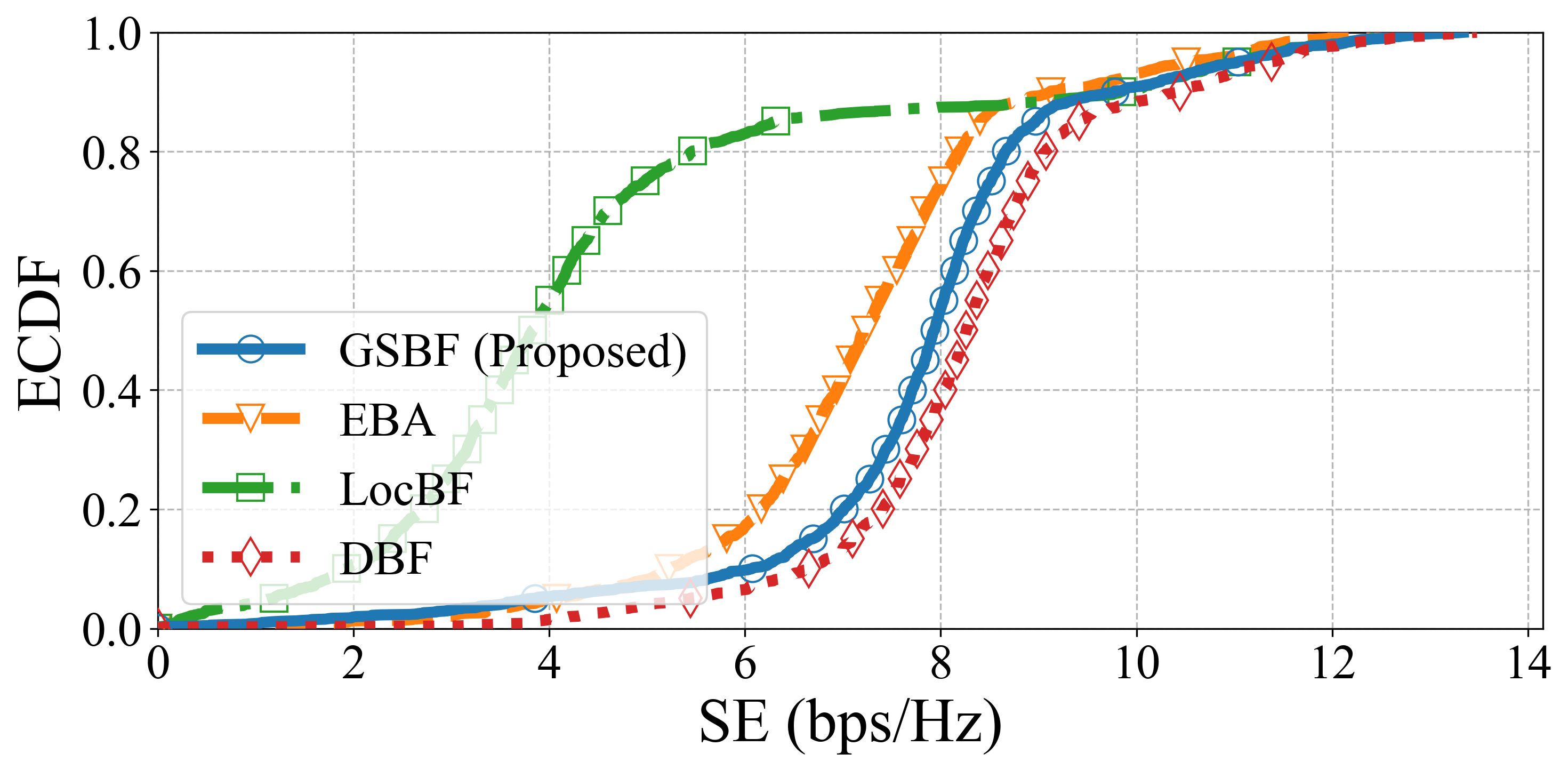}
    \caption{ECDF of SE with $ N_\text{x}=N_\text{y}=4 $.}
  \end{subfigure}
  \begin{subfigure}{0.323\textwidth}
    \centering
    \includegraphics[width=\linewidth]{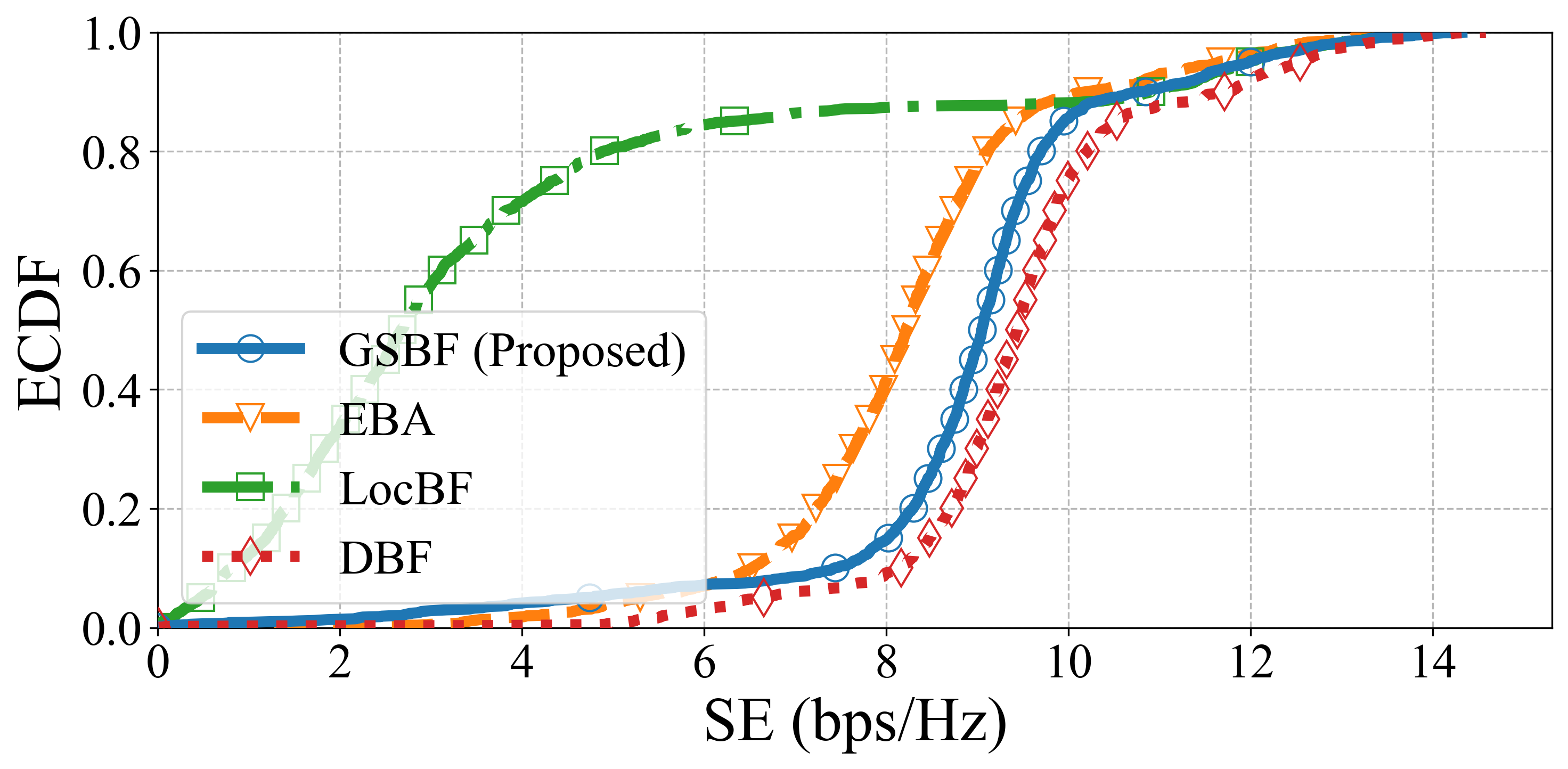}
    \caption{ECDF of SE with $ N_\text{x}=N_\text{y}=6 $.}
  \end{subfigure}
  \begin{subfigure}{0.323\textwidth}
    \centering
    \includegraphics[width=\linewidth]{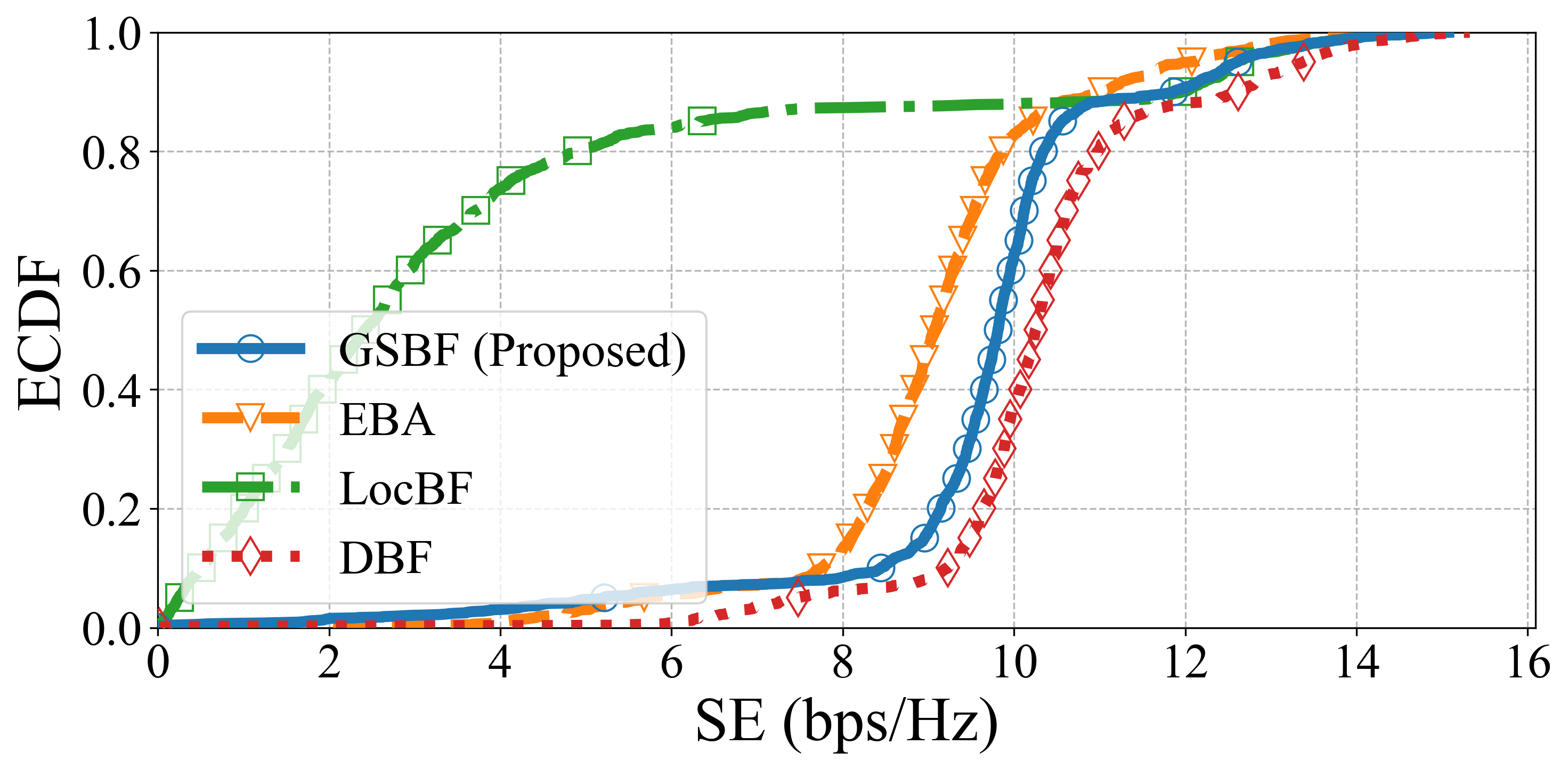}
    \caption{ECDF of SE with $ N_\text{x}=N_\text{y}=8 $.}
    \label{fig:ecdf_8x8}
  \end{subfigure}
  \caption{ECDF of SE performance versus different UPA size configurations.}
  \label{fig:results}
  % \vspace{-0.6cm}
\end{figure*}
\begin{table}[H]
  \centering
  \caption{Mean and Median SE performance.}
  \resizebox{\columnwidth}{!}{%
  \begin{tabular}{c|cccc}
    \toprule[1.5pt]
    \multirow{2}{*}{\textbf{Antenna Size}} & \multicolumn{4}{c}{\textbf{Mean / Median SE (bps/Hz)}} \\ \cline{2-5}
     & \textbf{GSBF (Ours)} & \textbf{EBA} & \textbf{LocBF} & \textbf{DBF (Upper Bound)} \\
    \midrule[1.5pt]
    $4 \times 4$ & \textbf{7.798 / 7.932} & 7.203 / 7.227 & 4.485 / 3.824 & 8.281 / 8.261 \\
    $6 \times 6$ & \textbf{8.874 / 9.051} & 8.238 / 8.216 & 3.771 / 2.671 & 9.490 / 9.437 \\
    $8 \times 8$ & \textbf{9.625 / 9.812} & 9.066 / 9.071 & 3.648 / 2.415 & 10.345 / 10.252 \\
    \bottomrule[1.5pt]
  \end{tabular}%
  }
  \label{tab:antenna_results_latest}
  % \vspace{-0.1cm}
\end{table}
We split the sampled poses into disjoint training and test sets and report mean and median values of SE compared to several baselines. The baselines include: (1) \textbf{DBF:} the fully digital architecture with the optimal beamformer, which is the upper bound of SE performance; (2) \textbf{EBA:} a discrete Fourier transform (DFT) codebook-based exhaustive beam search that always selects the beam with the largest virtual gain; (3) \textbf{LocBF:} a geometry-based beamforming baseline that computes the relative direction from AP position and orientation to the user position, then generates a beamforming vector from that direction at AP-side.

The experimental results for varying antenna scales are presented in Table \ref{tab:antenna_results_latest}. As the array size expands from $4\times4$ to $8\times8$, GSBF consistently outperforms both the EBA and LocBF baselines in both mean and median SE. GSBF maintains a consistent mean SE advantage over EBA across the considered array sizes. This superiority stems from the fact that GSBF enables fine-grained beam steering without the quantization losses inherent in the discrete codebooks used by EBA, and can effectively capture the complex scattering patterns in the indoor environment. Meanwhile, LocBF degrades as the array size increases due to its inability to account for NLoS multi-path components.

GSBF also achieves an average inference latency of 24.2 ms, 20.9 ms, and 23.5 ms for $4\times4$, $6\times6$, and $8\times8$ arrays, respectively. In contrast, EBA incurs substantial physical overhead from sequential beam sweeping: in practical mmWave systems \cite{tutorial_bm} with 8 synchronization signal blocks per 20 ms period, the alignment latency is approximately 40.0 ms, 100.0 ms, and 160.0 ms. While EBA's latency increases sharply with $N$, GSBF exhibits remarkable stability, since the rasterization complexity depends primarily on the number of Gaussian primitives and the panorama resolution, both largely independent of the array dimensions.

From the empirical cumulative distribution function (ECDF) in Fig. \ref{fig:results}, we further observe that the dataset mainly includes two distinct propagation regimes, which are inherently rooted in the heterogeneous propagation characteristics of the indoor factory environment. The ECDF curve of LocBF rises rapidly at first and then more gradually, indicating that NLoS paths dominate the evaluated region, while a subset of LoS paths remains. At high SE values, GSBF outperforms EBA and LocBF across all considered array sizes while remaining close to DBF, demonstrating its effectiveness in capturing complex multi-path effects and generating beams aligned with the dominant propagation paths. The remaining gap to the DBF upper bound results from both the constant-modulus analog beamforming constraint and the approximation error of the learned GSBF representation.
\section{Conclusion}
This paper proposed GSBF, a framework that synthesizes grid-free beamforming vectors directly from multi-modal data without online CSI. By designing panoramic projection, two-sided electromagnetic rasterization, and a reciprocity-preserving Bi-SG environment surrogate, GSBF renders an angular propagator map that is mapped to constant-modulus beams via an over-complete array-manifold dictionary. Simulations confirm that GSBF approaches the digital beamforming upper bound while outperforming baselines with millisecond-scale inference, establishing a new paradigm for environment-aware beamforming through the fusion of physical modeling and data-driven learning.

\normalem
%\cite{IEEEexample:articleetal}
\bibliographystyle{IEEEtran}
\bibliography{IEEEabrv,reference}

\newpage
\clearpage

\end{document}